\documentclass{article}
\usepackage{spconf,amsmath,graphicx,hyperref,amssymb,amsfonts}
\usepackage{multirow}
\usepackage{subfigure}
\usepackage{balance}

\title{LamiGauss: Pitching Radiative Gaussian for Sparse-View X-ray Laminography Reconstruction}
\name{\parbox{\linewidth}{\centering Chu Chen$^{1}$, Ander Biguri$^{2}$, Jean-Michel Morel$^{3}$, Raymond H. Chan$^{3,4}$, Carola-Bibiane Schönlieb$^{2}$,\\ and Jizhou Li$^{5}$}
\thanks{This work is partially funded by the National Natural Science Foundation of China (Nos. T2422017 and 52303301), the Hong Kong RGC (No. 21204124, CityU11309922 and LU13300125), the Shun Hing Institute of Advanced Engineering, CUHK (No. RNE-p1-25), ITF Grant (No. MHP/054/22, LU BGR 105824), and EPSRC grant EP/W004445/1.}
}
\address{
$^1$ Department of Mathematics, City University of Hong Kong, Hong Kong \\
$^2$ Department of Applied Mathematics and Theoretical Physics, University of Cambridge, UK\\
$^3$ School of Data Science, Lingnan University, Hong Kong \\
$^4$ Department of Operations and Risk Management, Lingnan University, Hong Kong\\
$^5$ Department of Electronic Engineering, The Chinese University of Hong Kong, Hong Kong }
\begin{document}
\ninept
\maketitle
\begin{abstract}
X-ray Computed Laminography (CL) is essential for non-destructive inspection of plate-like structures in applications such as microchips and composite battery materials, where traditional computed tomography (CT) struggles due to geometric constraints. However, reconstructing high-quality volumes from laminographic projections remains challenging, particularly under highly sparse-view acquisition conditions. In this paper, we propose a reconstruction algorithm, namely LamiGauss, that combines Gaussian Splatting radiative rasterization with a dedicated detector-to-world transformation model incorporating the laminographic tilt angle. LamiGauss leverages an initialization strategy that explicitly filters out common laminographic artifacts from the preliminary reconstruction, preventing redundant Gaussians from being allocated to false structures and thereby concentrating model capacity on representing the genuine object. Our approach effectively optimizes directly from sparse projections, enabling accurate and efficient reconstruction with limited data. Extensive experiments on both synthetic and real datasets demonstrate the effectiveness and superiority of the proposed method over existing techniques. LamiGauss uses only 3$\%$ of full views to achieve superior performance over the iterative method optimized on a full dataset.
\end{abstract}
\begin{keywords}
Computed Laminography, Gaussian Splatting, Sparse-view Reconstruction.
\end{keywords}
\section{Introduction}
\label{sec:intro}

X-ray Computed Tomography (CT) is a powerful imaging technique that enables the visualization of internal structures by reconstructing a 3D volume from a series of 2D X-ray projections from different angles. It has become an indispensable tool across a wide range of fields, including medical science~\cite{ritman2011current}, battery research~\cite{xue2022data}, biological research~\cite{rawson2020x}, and industrial inspection~\cite{stock1999x,stock2008recent}. However, conventional CT faces inherent limitations when inspecting flat, plate-like objects such as printed circuit boards, pouch battery cells, composite panels, or artworks.

\begin{figure*}[!ht]
\centering
  \includegraphics[width=1\linewidth]{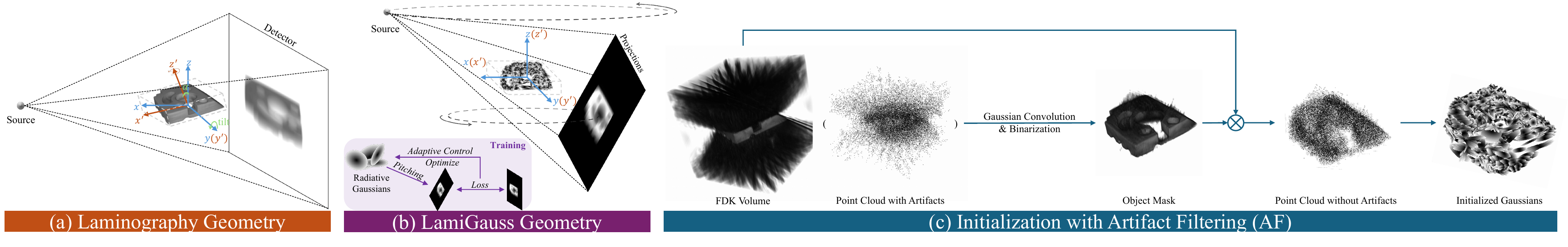}
  \caption{The schematic illustration of (a) laminography geometry, (b) the equivalent LamiGauss imaging geometry and the training process, and (c) the flowchart of its initialization strategy with artifact filtering.}
  \label{fig:laminogauss}
\end{figure*}

To address this challenge, Computed Laminography (CL) employs a specialized scanning geometry where the rotation axis of the sample is tilted at an angle of less than $90^\circ$ to the horizontal \cite{helfen2005high,o2016recent} as shown in Fig.~\ref{fig:laminogauss}(a). This configuration ensures that for any given rotation angle, the X-ray beam traverses a relatively short path through the plate-like object, significantly improving penetration and signal-to-noise ratio. This capability has enabled its successful application across various fields, including ptycho-laminography imaging of integrated circuits~\cite{kang2023accelerated}, in situ monitoring of battery component evolution~\cite{zan2022situ}, and high-resolution inspection of large-sized samples~\cite{nikitin2023tomocupy,nikitin2024laminography}. Despite its advantageous geometry, high-quality CL reconstruction remains a challenging inverse problem, particularly under highly sparse-view acquisition settings aimed at reducing scanning time. Traditional analytical algorithms, such as the FDK algorithm~\cite{feldkamp1984practical}, are computationally efficient but produce severe streak artifacts when projections are limited. Iterative algorithms ~\cite{fista,fistatv,cgls} incorporate prior knowledge (e.g., total variation regularization) to suppress noise and artifacts but often suffer from over-smoothed textures and prolonged computation time.

Recently, 3D Gaussian Splatting (GS) \cite{kerbl3Dgaussians} has emerged as a groundbreaking technique in computer vision for novel view synthesis. It represents a 3D scene with a set of anisotropic Gaussians and utilizes a highly efficient differentiable rasterizer for rendering, demonstrating remarkable rendering quality and speed. This paradigm has been extended to X-ray imaging \cite{cai2024radiative,zha2024r2,gao2024ddgs}, with impressive performance for synthesizing novel views. However, being a generalization of CT geometry, these methods lack geometry flexibility and fail in CL representation.


We propose LamiGauss, a framework for direct, high-quality CL reconstruction from sparse projections. Our method introduces a rasterization process with CL geometry and an artifact filtering (AF) initialization that suppresses common laminographic artifacts, enabling faster convergence and better reconstructions. Experiments on synthetic and real datasets show that LamiGauss outperforms baselines in both accuracy and efficiency, even surpassing traditional methods trained on thousands of views using only 80.

The main contributions are summarized as:
(1) We propose LamiGauss, the first 3DGS-based framework for direct CL reconstruction, which integrates the transformation for the laminographic tilt angle; (2) We introduce the artifact filtering (AF) initialization that effectively prevents the wasteful allocation of Gaussians on laminographic artifact regions;
(3) Extensive experiments on synthetic and real data demonstrate that our method achieves superior reconstruction quality, especially in highly sparse-view settings.

\section{Method}
\label{sec:method}

Our goal is to reconstruct the 3D attenuation coefficient distribution $\sigma(\mathbf{x})$ of an object from a highly sparse set of its X-ray laminography projections $\{\mathbf{I}_i\}_{i=1,\cdots, N}$. The LamiGauss consists of four key components: 1) initialization with artifact filtering (AF), 2) radiative Gaussians representation, 3) X-ray rasterization process under laminography geometry, and 4) optimization. An overview of our pipeline is illustrated in Fig.~\ref{fig:laminogauss}.

\subsection{Radiative Gaussians Representation}
\label{ssec:representation}

We represent the scanned object as a set of learnable 3D Gaussian kernels $\mathbb{G}^3 = \{G_{i}^{3}\}_{i=1,\cdots,M}$, termed \emph{radiative Gaussians} follow~\cite{cai2024radiative,zha2024r2}. Each kernel $G_{i}^{3}$ defines a local Gaussian-shaped density field:
\begin{equation}
G_{i}^{3}(\mathbf{x}|\rho_{i},\mathbf{p}_{i},\boldsymbol{\Sigma}_{i}) = \rho_{i} \cdot \exp\left(-\frac{1}{2}(\mathbf{x}-\mathbf{p}_{i})^{\top}\boldsymbol{\Sigma}_{i}^{-1}(\mathbf{x}-\mathbf{p}_{i})\right),
\end{equation}
where $\rho_{i} \in \mathbb{R}^{+}$, $\mathbf{p}_{i} \in \mathbb{R}^{3}$, and $\boldsymbol{\Sigma}_{i} \in \mathbb{R}^{3\times3}$ are the learnable parameters representing the central density, position, and covariance matrix of the $i$-th Gaussian, respectively. The overall density at any point $\mathbf{x} \in \mathbb{R}^{3}$ is computed by summing the contributions from all kernels:
\begin{equation}
\sigma(\mathbf{x}) = \sum_{i=1}^{M} G_{i}^{3}(\mathbf{x}|\rho_{i},\mathbf{p}_{i},\boldsymbol{\Sigma}_{i}).
\label{eq:density}
\end{equation}
For stability~\cite{kerbl3Dgaussians}, the covariance matrix $\boldsymbol{\Sigma}_{i}$ is parameterized by a rotation matrix $\mathbf{R}_{i}$ and a scaling matrix $\mathbf{S}_{i}$: $\boldsymbol{\Sigma}_{i} = \mathbf{R}_{i}\mathbf{S}_{i}\mathbf{S}_{i}^{\top}\mathbf{R}_{i}^{\top}$.

\subsection{X-ray Rasterization under Laminography Geometry}
\label{ssec:rasterization}

In X-ray laminography, a projection image $\mathbf{I}$ captures the line integrals of the attenuation coefficients. For a ray $\mathbf{r}(t) = \mathbf{o} + t\mathbf{d}$ with near and far bounds $t_n$ and $t_f$, the pixel value in the logarithmic domain is given by the Beer-Lambert law:
\begin{equation}
I(\mathbf{r}) = \log\left(\frac{I_0}{I'(\mathbf{r})}\right) = \int_{t_n}^{t_f} \sigma(\mathbf{r}(t)) \, dt,
\label{eq:xray_form}
\end{equation}
where $I_0$ is the initial intensity and $I'(\mathbf{r})$ is the detected intensity. Substituting Eq.~(\ref{eq:density}) into Eq.~(\ref{eq:xray_form}) yields:
\begin{equation}
I(\mathbf{r}) = \int \sum_{i=1}^{M} G^{3}_{i}(\mathbf{r}(t)) \, dt = \sum_{i=1}^{M} \int G^{3}_{i}(\mathbf{r}(t)) \, dt.
\label{eq:xray_gauss}
\end{equation}
This allows us to render the projection by individually integrating each 3D Gaussian along the ray.

The core of our Laminography adaptation lies in the geometric transformation from world space to ray space. Unlike standard cone-beam CT, which rotates around a vertical axis, laminography employs a tilted rotation axis with respect to the incoming beam, the sample surface normal being approximately parallel to the rotation axis \cite{helfen2005high,xu2012comparison}. 


By introducing the tilt angle $\alpha$ that defines the fixed inclination between the rotation axis ($z'$) and the detector plane, the viewing transformation matrix $\mathbf{T}_{\text{CL}}$ for laminography geometry becomes:
\begin{align}
\mathbf{T}_{\text{CL}} &= \begin{bmatrix}
\mathbf{W}_{\text{CL}} & \mathbf{t}_{\text{CL}} \\
\mathbf{0} & 1
\end{bmatrix}, \quad \mathbf{t}_{\text{CL}} = D_{SO} \begin{bmatrix}
-\cos\theta\sin\alpha \\ -\sin\theta\sin\alpha \\ \cos\alpha
\end{bmatrix}, \\
\mathbf{W}_{\text{CL}} &= \begin{bmatrix}
-\sin\theta & \cos\theta & 0 \\
\cos\theta\sin\alpha & \sin\theta\sin\alpha & -\cos\alpha \\
-\cos\theta\cos\alpha & -\sin\theta\cos\alpha & -\sin\alpha
\end{bmatrix}. \label{eq:lamtrans}
\end{align}
where $D_{SO}$ is the source-to-object distance. Thus, the transformation matrix in CT~\cite{zha2024r2} is the trivial case of Eq.~(\ref{eq:lamtrans}) when $\alpha = 0$. A point $\mathbf{p}$ and its covariance $\boldsymbol{\Sigma}$ in the world space are transformed into the ray space $\tilde{\mathbf{p}}$ and $\tilde{\boldsymbol{\Sigma}}$ using: $\tilde{\mathbf{p}} = \phi(\mathbf{p})$, and $ \tilde{\boldsymbol{\Sigma}} = \mathbf{J} \mathbf{W}_{\text{CL}} \boldsymbol{\Sigma} \mathbf{W}_{\text{CL}}^{\top} \mathbf{J}^{\top}$, where $\phi$ is the projective mapping and $\mathbf{J}$ is the Jacobian of the local affine approximation \cite{zwicker2002ewa}. The geometry of LamiGauss during training is visualized in Fig.~\ref{fig:laminogauss}(b).

Following the transformation, a normalized 3D Gaussian integrated along the ray direction yields a normalized 2D Gaussian on the detector plane. With the integral of the $i$-th Gaussian being $
\int G^{3}_{i}(\mathbf{r}(t)) \, dt \approx G^{2}_{i}(\hat{\mathbf{x}}|\hat{\rho}_{i}, \hat{\mathbf{p}}_{i}, \hat{\boldsymbol{\Sigma}}_{i})$ and Eq.~(\ref{eq:xray_gauss}), the final pixel value is rendered by summing the contributions of all 2D Gaussians:
\begin{equation}
I_{r}(\mathbf{r}) = \sum_{i=1}^{M} G^{2}_{i}(\hat{\mathbf{x}}|\hat{\rho}_{i}, \hat{\mathbf{p}}_{i}, \hat{\boldsymbol{\Sigma}}_{i}).
\end{equation}
where $\hat{\mathbf{p}}_{i}$ and $\hat{\boldsymbol{\Sigma}}_{i}$ are the 2D position and covariance by dropping the third component of $\tilde{\mathbf{p}}_{i}$ and the third row and column of $\tilde{\boldsymbol{\Sigma}}_{i}$, respectively, and $\hat{\rho}_{i} = \mu_i \rho_i$, with $\mu_i = \sqrt{2\pi {|\tilde{\boldsymbol{\Sigma}}_{i}|}/{|\hat{\boldsymbol{\Sigma}}_{i}|}}$ to avoid integration bias~\cite{zha2024r2}.

\begin{figure*}
\centering
  \includegraphics[width=1\linewidth]{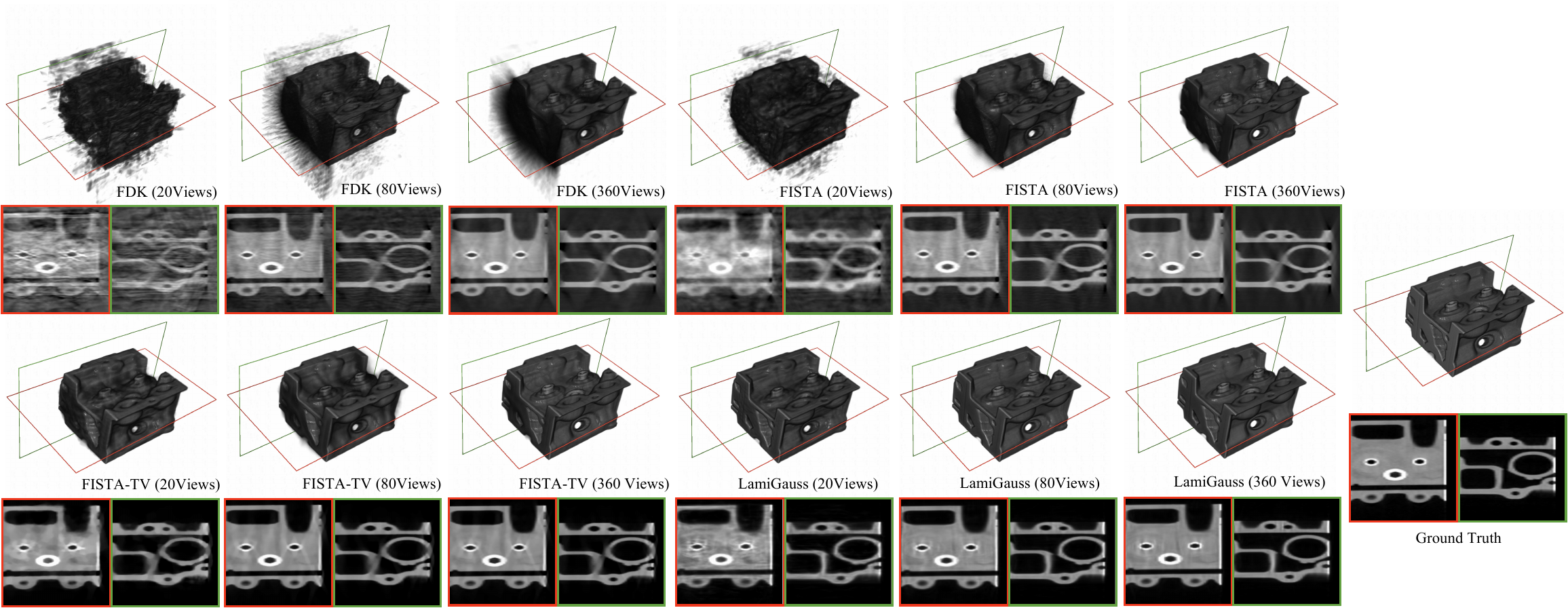}
  \caption{Visualization of reconstruction results on the synthetic \emph{Engine} under different sparse-view conditions. 
  }
  \label{fig:simvis}
\end{figure*}

\begin{figure}[ht]
\centering
\vspace{-5pt}
  \includegraphics[width=1\linewidth]{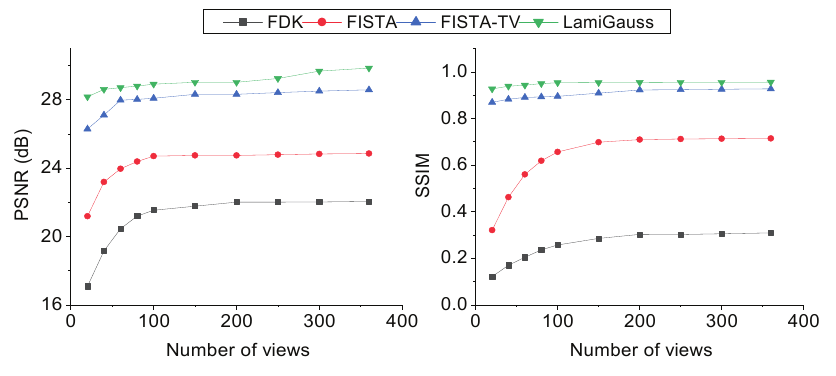}
  \caption{Quantitative evaluation of different reconstruction methods on the synthetic \emph{Engine} with respect to the number of reference projections, where red represents SSIM and blue PSNR. Our proposed LamiGauss consistently outperforms all baselines.
  }
  \label{fig:quant}
  \vspace{-10pt}
\end{figure}

\subsection{Initialization with Artifact Filtering}
\label{ssec:init}

A high-quality initialization is crucial for the convergence speed and final performance of Gaussian-based optimization. A naïve approach would be to use FDK, but in laminography this would cause severe artifacts in the image (see Fig.~\ref{fig:laminogauss}(c)) that would introduce a large portion of wrong primitives. Iterative algorithms would take too much time. To address this, we propose an initialization strategy with explicit AF. Our goal is to generate a sparse point cloud that concentrates primarily on the genuine object, thereby guiding the subsequent optimization to focus its resources on meaningful areas. The process, illustrated in Fig.~\ref{fig:laminogauss}(c), consists of the following steps:


1) The FDK-reconstructed volume $\mathbf{V}_{\text{FDK}}$ is first convolved with a 3D isotropic Gaussian kernel $\mathcal{K}$ to supress the high frequency laminograpgic artifacts.

2) The smoothed volume is then binarized using Otsu's method \cite{otsu1975threshold}, which automatically determines an optimal threshold value $\tau$ to separate the foreground object from the background and residual artifacts, and results in a 3D mask $\mathbf{M}$ that approximately delineates the object's domain:

\vspace{-5pt}
\begin{equation}
\mathbf{M}(\mathbf{x}) = 
\begin{cases} 
1 & \text{if } \mathcal{K} * \mathbf{V}_{\text{FDK}} \geq \tau \\
0 & \text{otherwise}
\end{cases}.
\vspace{-2pt}
\end{equation}

3) Finally, $M$ points are randomly sampled from masked region as $\{\mathbf{x}_i\}_{i=1}^M = \mathcal{S}(\{ \mathbf{x} | \mathbf{M}(\mathbf{x})=1\})$,
where $\mathcal{S}$ is random sampling operator. The initial central density is queried from the FDK volume as $\rho_i = \mathbf{V}_{\text{FDK}}(\mathbf{x}_i)$.

This tailored initialization procedure ensures the limited number of Gaussians is allocated efficiently from the very start of training. 


\subsection{Optimization}
\label{ssec:optimization}

We optimize the parameters of the radiative Gaussians $\Theta = \{\rho_i, \mathbf{p}_i, \boldsymbol{\Sigma}_i\}$ by minimizing the 2D rendering loss:
\begin{equation}
\mathcal{L}_{\text{total}} = \mathcal{L}_{1}(I_{r}, I_{m}) + \lambda_{\text{ssim}} \left(1- \mathcal{L}_{\text{ssim}}(I_{r}, I_{m})\right)
\end{equation}
Here, $I_{r}$ and $I_{m}$ are the rendered and measured projections, respectively. $\mathcal{L}_{1}$ and $\mathcal{L}_{\text{ssim}}$ are the $\ell_1$ and SSIM~\cite{wang2004image} losses applied to the 2D images. An adaptive density control strategy \cite{zha2024r2} is used to clone, split, or prune Gaussians based on the gradient of the photometric loss to better represent the scene.


\begin{figure}[!t]
\centering
    \subfigure[FDK]{
{\includegraphics[width=0.23\linewidth]{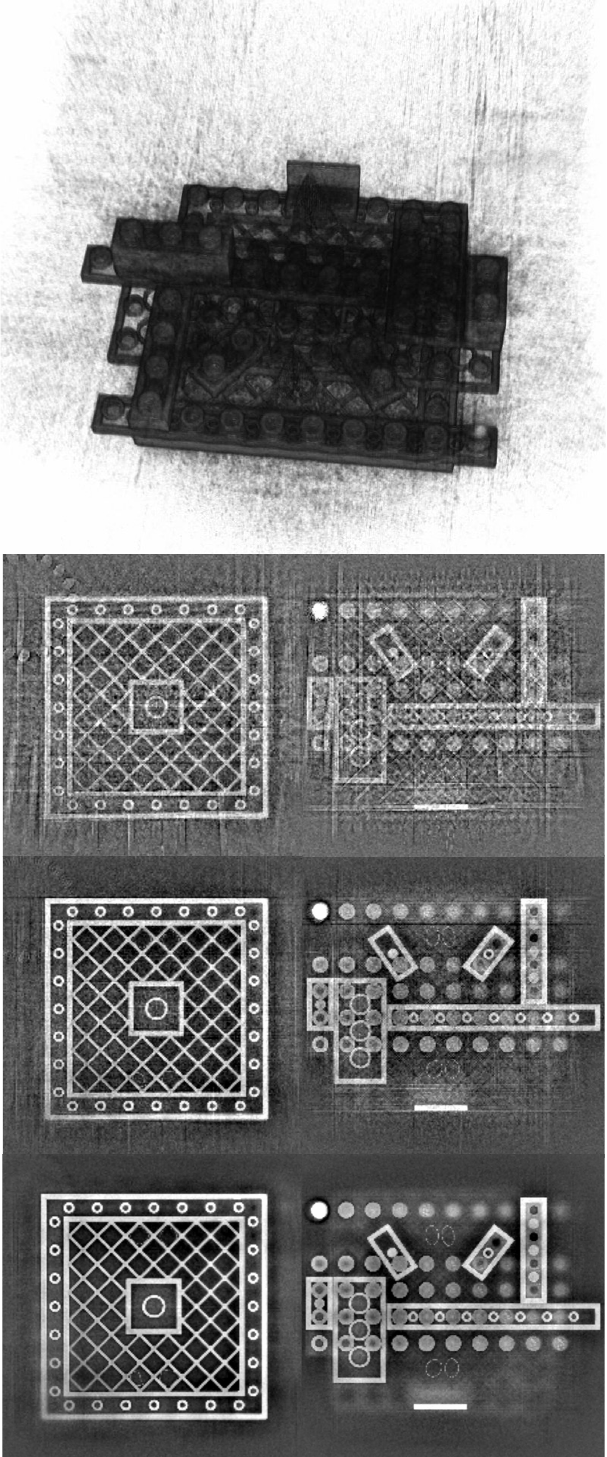}}}
   \subfigure[FISTA]{
{\includegraphics[width=0.23\linewidth]{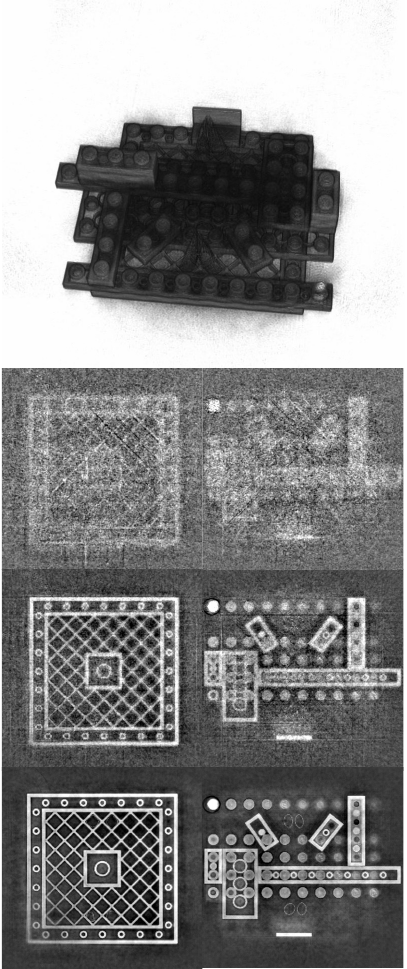}}}
    \subfigure[FISTA-TV]{
{\includegraphics[width=0.23\linewidth]{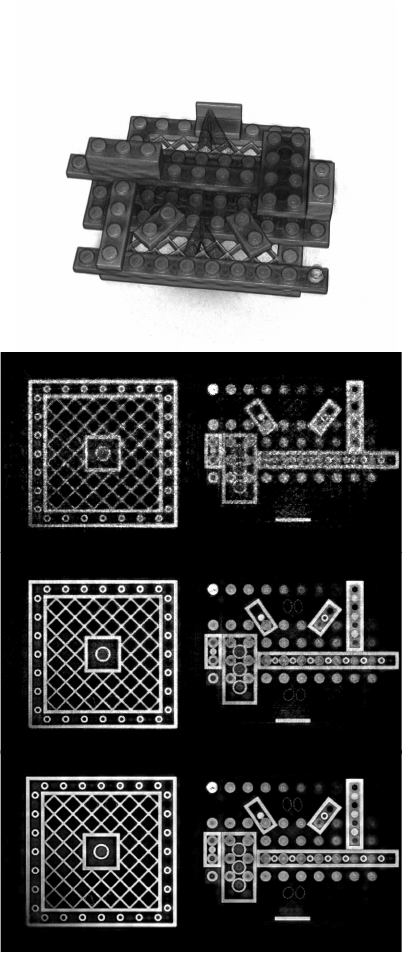}}}
    \subfigure[LamiGauss]{
{\includegraphics[width=0.23\linewidth]{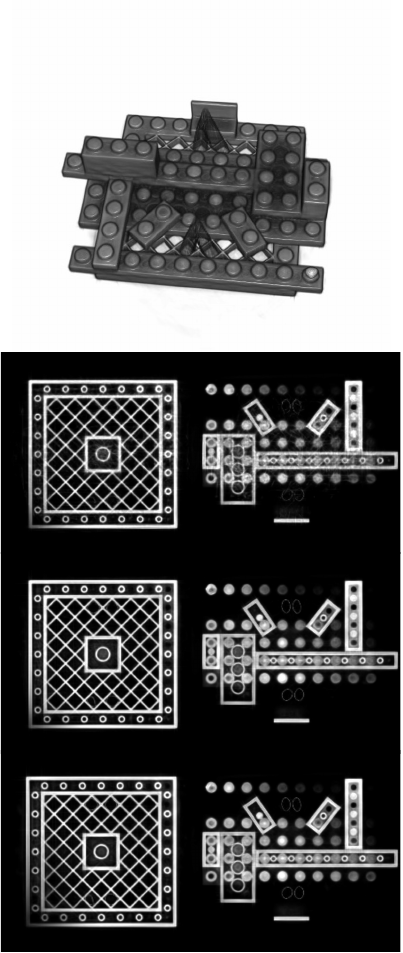}}}

\vspace{-5pt}
    \caption{Visual comparison from real laminography projections. The first row displays the reconstructed \emph{LEGO} of various methods from 80 views, where two axial slices obtained from 20/80/360 projections are shown in the following 2nd/3rd/4th row, respectively.}
    \label{fig:legovis}
\end{figure}

\begin{figure}[!ht]
\centering
    \subfigure[CGLS (2513 views)]{
{\includegraphics[width=0.8\linewidth]{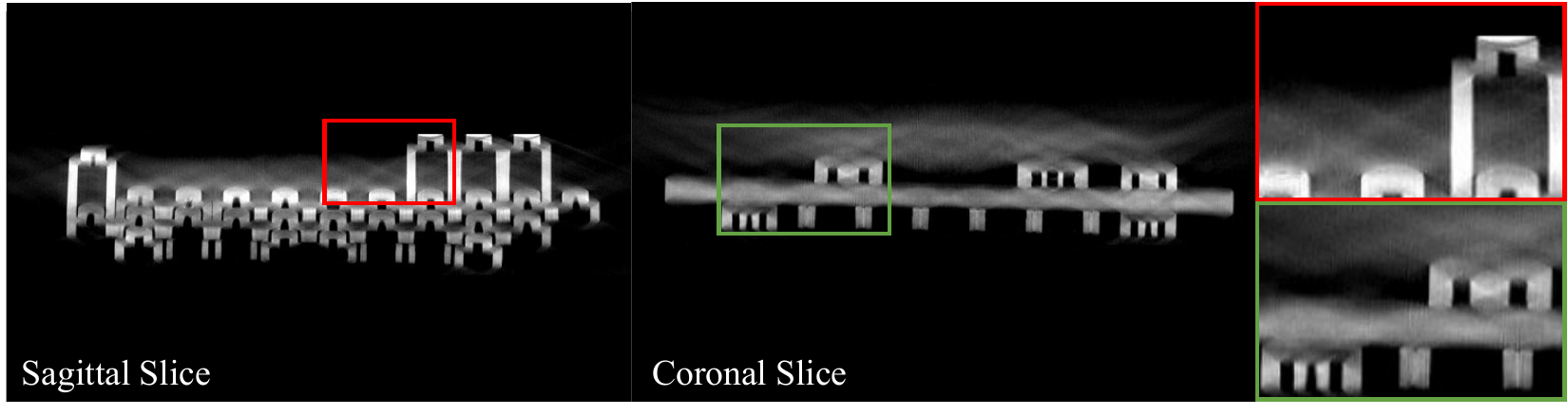}}}
   \subfigure[LamiGauss (80 views)]{
{\includegraphics[width=0.8\linewidth]{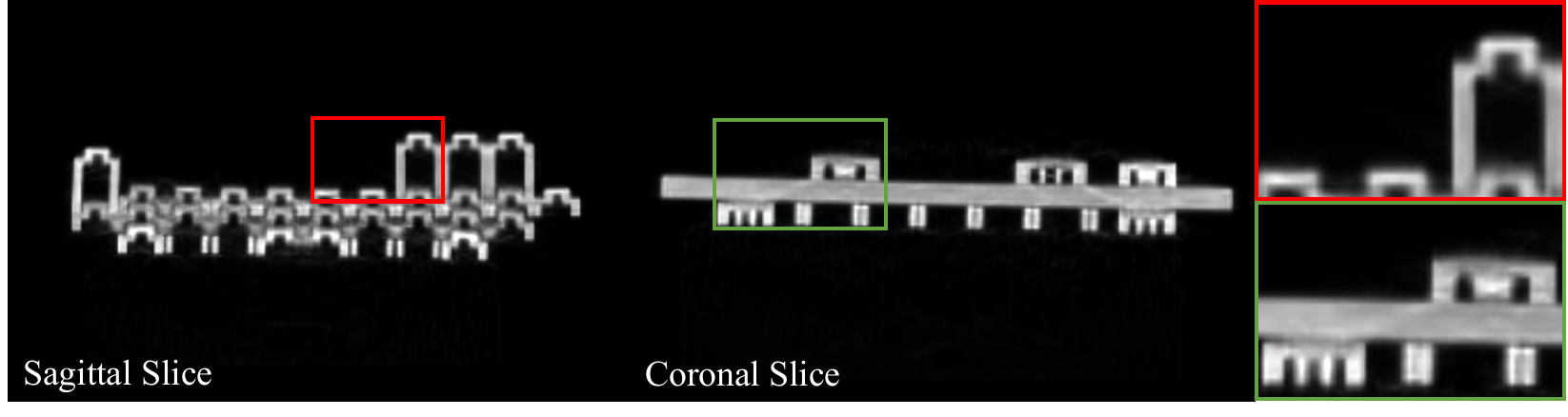}}}
    \caption{Laminography reconstruction by (a) CGLS on 2513 projections and (b) LamiGauss on 80 projections. The details in the region of the bounding boxes are zoomed in on the right, demonstrating the capability of LamiGauss in removing laminographic artifacts.}
    \label{fig:lego_det}
    \vspace{-10pt}
\end{figure}

\section{Experiments}
\label{sec:experiment}

\subsection{Experimental Setup}

\textbf{Dataset} To comprehensively evaluate the performance of our proposed LamiGauss, we conduct experiments on both synthetic and real-world laminography datasets. For synthetic evaluation, we use the \textit{Engine} model from the SciVis dataset\cite{scivisdata}, simulating a $30^\circ$ tilt angle acquisition with $512 \times 512$ projections over $360^\circ$ using CIL~\cite{jorgensen2021core}. For real data, we use the \emph{LEGO brick}~\cite{fisher2019laminography}, which contains 2513 projections ($1148 \times 1596$) acquired at $\alpha = 30^\circ$, along with a CGLS~\cite{cgls} reconstruction from all views as reference. We visualize sparse-view performance with 20, 80, and 360 uniformly subsampled projections to analyze the method's robustness under extreme to moderate sparsity.


\textbf{Implementation Details} All methods, FDK~\cite{feldkamp1984practical}, FISTA~\cite{fista}, FISTA-TV~\cite{fistatv}, and our LamiGauss, are implemented in PyTorch, leveraging the TIGRE toolbox~\cite{Biguri2016tigre, Biguri2025tigrev3} and executed on an NVIDIA RTX 3090 GPU.  We train our model for 30k iterations using Adam optimizer, initialize with $M = 30\text{k}$ Gaussians (Sec.~\ref{ssec:init}), use $\lambda_{\text{ssim}} = 0.25$, and apply adaptive control from iterations 500 to 20k with a gradient threshold of $5\times10^{-5}$. Performance evaluation is based on PSNR (for the 3D volume) and SSIM (averaged over 2D slices).

\subsection{Simulations}
\vspace{-2pt}

We first evaluate the reconstruction performance of our proposed LamiGauss from synthetic projections. As shown in Fig.~\ref{fig:simvis}, classical methods struggle significantly under the inherent challenges of laminography geometry and high sparsity. The analytical FDK algorithm produces severe striped artifacts that obscure most structural details. While iterative methods like FISTA and FISTA-TV improve upon FDK by reducing noise, they still result in feature loss and adjacent structures blurred or fused. These effects are particularly pronounced in objects with internal cavities and complex piping, and persist even as the number of views increases to 360. 

In contrast, LamiGauss produces clear and sharp reconstructions across all sparsity levels, with the complex internal architecture of the engine resolved, maintaining clear separations between closely spaced structures and preserving sharp edges without fusing. This capability stems from the powerful representation capacity of the Gaussians and our effective AF initialization.

The quantitative results in Fig.~\ref{fig:quant} further support the visual observations. LamiGauss achieves performance across all tested numbers of views, whereas other baselines exhibit dramatic degradation as the number of projections is reduced. The significant gap in the extreme sparse-view condition underscores the method's robustness and efficiency in learning from very limited data.

\subsection{Real Acquisition}

The superior performance of LamiGauss is further validated on the real-world \emph{LEGO} dataset. As clearly visible in the axial slices in Fig.~\ref{fig:legovis}, LamiGauss reconstruction using only 20 views already surpasses the quality of both FDK and FISTA reconstructions that utilized 360 views. While the FISTA-TV method shows improved noise suppression over FISTA, its results at 20 and 80 views are still contaminated by substantial speckle texture, failing to recover clean structural boundaries. 


The most compelling evidence for the capability of our method is revealed in the sagittal and coronal planes (Fig.~\ref{fig:lego_det}). These views expose the persistent, blurry strip-like artifacts above the object, which are characteristic of the laminography geometry and remain evident even in the CGLS reconstruction from full projections. Remarkably, LamiGauss reconstruction, trained on a mere $3\%$ of the projections (80 views), is completely free of these inherent laminography artifacts. This demonstrates that our method does not merely approximate the ``gold standard'' but effectively learns a superior, artifact-free representation of the object. Furthermore, this breakthrough in quality is achieved with unprecedented efficiency. While the CGLS method required several hours to converge on the full dataset with GPU acceleration, our model produces a superior result in ~10 minutes. This combination of exceptional accuracy, robustness to extreme sparsity, and computational efficiency suggests LamiGauss is well-suited for the X-ray laminography reconstruction task.

\subsection{Ablation Study}

We evaluate the effectiveness of the artifact filtering (AF) initialization. The quantitative results are summarized in Table~\ref{tab:ablation}. The comparison clearly demonstrates that employing AF brings significant benefits to the LamiGauss framework, as it not only achieves a superior final reconstruction quality at the same number of iterations but also exhibits dramatically improved convergence speed. 


\vspace{-5pt}
\begin{table}[htbp]
\centering
\caption{Ablation results with or without artifact filtering.}
\begin{tabular}{c c c c c}
\hline
        Methods    & \# Iters   &  PSNR (dB) $\uparrow$ &  SSIM $\uparrow$ & Time (s) \\
\hline
\multirow{3}{*}{w/o AF} & 1k & 19.9208 & 0.7489 & 50 \\
                        & 2k & 21.8731 & 0.8189 & 118 \\
                        & 5k & 22.6186 & 0.8634 & 254 \\
\hline
\multirow{3}{*}{w AF}   & 1k & 20.6113 & 0.7638 & 32 \\
                        & 2k & 22.5023 & 0.8213 & 67 \\
                        & 5k & 24.1821 & 0.8954 & 183 \\
\hline
\end{tabular}\\
\label{tab:ablation}
{\footnotesize *Ablation studies conducted on \emph{Engine} object from 50 views.}
\end{table}
\vspace{-10pt}

\section{Conclusions}
In this paper, we present LamiGauss, a highly effective reconstruction framework for X-ray Computed Laminography (CL), which is the first 3D Gaussian Splatting-based method to our knowledge. Our key innovation lies in the integration of the CL imaging geometry into a differentiable rendering pipeline, coupled with the artifact filtering (AF) initialization, which proactively filters out common laminographic artifacts from the initial point cloud. This significantly enhances optimization efficiency and leads to reconstructions that outperform in real data full scan with only extremely sparse sampling, with a fraction of the computational time. This superior outcome highlights an unprecedented trade-off between computational cost, reconstruction fidelity, and laminographic artifacts removal. By surpassing the long-standing gold standard, our work establishes a new paradigm for high-quality and efficient X-ray CL imaging, with strong potential to impact real-world applications.

\balance
\bibliographystyle{IEEEbib}
\bibliography{strings,ref}

\begin{thebibliography}{10}

\bibitem{ritman2011current}
Erik~L Ritman,
\newblock ``Current status of developments and applications of micro-{CT},''
\newblock {\em Annual review of biomedical engineering}, vol. 13, no. 1, pp. 531--552, 2011.

\bibitem{xue2022data}
Zhichen Xue, Jizhou Li, Piero Pianetta, and Yijin Liu,
\newblock ``Data-driven lithium-ion battery cathode research with state-of-the-art synchrotron {X}-ray techniques,''
\newblock {\em Accounts of Materials Research}, vol. 3, no. 8, pp. 854--865, 2022.

\bibitem{rawson2020x}
Shelley~D Rawson, Jekaterina Maksimcuka, Philip~J Withers, and Sarah~H Cartmell,
\newblock ``X-ray computed tomography in life sciences,''
\newblock {\em BMC biology}, vol. 18, no. 1, pp. 21, 2020.

\bibitem{stock1999x}
SR~Stock,
\newblock ``X-ray microtomography of materials,''
\newblock {\em International materials reviews}, vol. 44, no. 4, pp. 141--164, 1999.

\bibitem{stock2008recent}
SR~Stock,
\newblock ``Recent advances in {X-ray} microtomography applied to materials,''
\newblock {\em International materials reviews}, vol. 53, no. 3, pp. 129--181, 2008.

\bibitem{helfen2005high}
L~Helfen, T~Baumbach, Petr Mikulik, D~Kiel, P~Pernot, P~Cloetens, and J~Baruchel,
\newblock ``High-resolution three-dimensional imaging of flat objects by synchrotron-radiation computed laminography,''
\newblock {\em Applied Physics Letters}, vol. 86, no. 7, 2005.

\bibitem{o2016recent}
Neil~S O’Brien, Richard~P Boardman, Ian Sinclair, and Thomas Blumensath,
\newblock ``Recent advances in {X-ray} cone-beam computed laminography,''
\newblock {\em Journal of X-ray Science and Technology}, vol. 24, no. 5, pp. 691--707, 2016.

\bibitem{kang2023accelerated}
Iksung Kang, Yi~Jiang, Mirko Holler, Manuel Guizar-Sicairos, Anthony~FJ Levi, Jeffrey Klug, Stefan Vogt, and George Barbastathis,
\newblock ``Accelerated deep self-supervised ptycho-laminography for three-dimensional nanoscale imaging of integrated circuits,''
\newblock {\em Optica}, vol. 10, no. 8, pp. 1000--1008, 2023.

\bibitem{zan2022situ}
Guibin Zan, Guannan Qian, Sheraz Gul, Jizhou Li, Katie Matusik, Yong Wang, Sylvia Lewis, Wenbing Yun, Piero Pianetta, David~J Vine, et~al.,
\newblock ``In situ visualization of multicomponents coevolution in a battery pouch cell,''
\newblock {\em Proceedings of the National Academy of Sciences}, vol. 119, no. 29, pp. e2203199119, 2022.

\bibitem{nikitin2023tomocupy}
Viktor Nikitin,
\newblock ``Tomocupy--efficient gpu-based tomographic reconstruction with asynchronous data processing,''
\newblock {\em Synchrotron Radiation}, vol. 30, no. 1, pp. 179--191, 2023.

\bibitem{nikitin2024laminography}
Viktor Nikitin, Gregg Wildenberg, Alberto Mittone, Pavel Shevchenko, Alex Deriy, and Francesco De~Carlo,
\newblock ``Laminography as a tool for imaging large-size samples with high resolution,''
\newblock {\em Synchrotron Radiation}, vol. 31, no. 4, pp. 851--866, 2024.

\bibitem{feldkamp1984practical}
Lee~A Feldkamp, Lloyd~C Davis, and James~W Kress,
\newblock ``Practical cone-beam algorithm,''
\newblock {\em Journal of the Optical Society of America A}, vol. 1, no. 6, pp. 612--619, 1984.

\bibitem{fista}
Amir Beck and Marc Teboulle,
\newblock ``A fast iterative shrinkage-thresholding algorithm for linear inverse problems,''
\newblock {\em SIAM journal on imaging sciences}, vol. 2, no. 1, pp. 183--202, 2009.

\bibitem{fistatv}
Amir Beck and Marc Teboulle,
\newblock ``Fast gradient-based algorithms for constrained total variation image denoising and deblurring problems,''
\newblock {\em IEEE transactions on image processing}, vol. 18, no. 11, pp. 2419--2434, 2009.

\bibitem{cgls}
Magnus~R Hestenes, Eduard Stiefel, et~al.,
\newblock ``Methods of conjugate gradients for solving linear systems,''
\newblock {\em Journal of research of the National Bureau of Standards}, vol. 49, no. 6, pp. 409--436, 1952.

\bibitem{kerbl3Dgaussians}
Bernhard Kerbl, Georgios Kopanas, Thomas Leimk{\"u}hler, and George Drettakis,
\newblock ``{3D} gaussian splatting for real-time radiance field rendering,''
\newblock {\em ACM Transactions on Graphics}, vol. 42, no. 4, July 2023.

\bibitem{cai2024radiative}
Yuanhao Cai, Yixun Liang, Jiahao Wang, Angtian Wang, Yulun Zhang, Xiaokang Yang, Zongwei Zhou, and Alan Yuille,
\newblock ``Radiative gaussian splatting for efficient x-ray novel view synthesis,''
\newblock in {\em European Conference on Computer Vision}. Springer, 2024, pp. 283--299.

\bibitem{zha2024r2}
Ruyi Zha, Tao~Jun Lin, Yuanhao Cai, Jiwen Cao, Yanhao Zhang, and Hongdong Li,
\newblock ``{R2-Gaussian}: Rectifying radiative gaussian splatting for tomographic reconstruction,''
\newblock {\em Advances in Neural Information Processing Systems}, vol. 37, pp. 44907--44934, 2024.

\bibitem{gao2024ddgs}
Zhongpai Gao, Benjamin Planche, Meng Zheng, Xiao Chen, Terrence Chen, and Ziyan Wu,
\newblock ``{DDGS-CT}: Direction-disentangled gaussian splatting for realistic volume rendering,''
\newblock {\em Advances in Neural Information Processing Systems}, vol. 37, pp. 39281--39302, 2024.

\bibitem{xu2012comparison}
Feng Xu, Lukas Helfen, Tilo Baumbach, and Heikki Suhonen,
\newblock ``Comparison of image quality in computed laminography and tomography,''
\newblock {\em Optics Express}, vol. 20, no. 2, pp. 794--806, 2012.

\bibitem{zwicker2002ewa}
Matthias Zwicker, Hanspeter Pfister, Jeroen Van~Baar, and Markus Gross,
\newblock ``{EWA} splatting,''
\newblock {\em IEEE Transactions on Visualization and Computer Graphics}, vol. 8, no. 3, pp. 223--238, 2002.

\bibitem{otsu1975threshold}
Nobuyuki Otsu et~al.,
\newblock ``A threshold selection method from gray-level histograms,''
\newblock {\em Automatica}, vol. 11, no. 285-296, pp. 23--27, 1975.

\bibitem{wang2004image}
Zhou Wang, Alan~C Bovik, Hamid~R Sheikh, and Eero~P Simoncelli,
\newblock ``Image quality assessment: from error visibility to structural similarity,''
\newblock {\em IEEE transactions on image processing}, vol. 13, no. 4, pp. 600--612, 2004.

\bibitem{scivisdata}
Pavol Klacansky,
\newblock ``Open {SciVis} datasets,'' December 2017,
\newblock \small \texttt{https://klacansky.com/open-scivis-datasets/}.

\bibitem{jorgensen2021core}
Jakob~S J{\o}rgensen, Evelina Ametova, Genoveva Burca, Gemma Fardell, Evangelos Papoutsellis, Edoardo Pasca, Kris Thielemans, Martin Turner, Ryan Warr, William~RB Lionheart, et~al.,
\newblock ``{Core Imaging Library-Part} {I}: a versatile python framework for tomographic imaging,''
\newblock {\em Philosophical Transactions of the Royal Society A}, vol. 379, no. 2204, pp. 20200192, 2021.

\bibitem{fisher2019laminography}
Sarah~L Fisher, DJ~Holmes, Jakob~Sauer J{\o}rgensen, Parmesh Gajjar, Julia Behnsen, William~RB Lionheart, and Philip~J Withers,
\newblock ``Laminography in the lab: imaging planar objects using a conventional x-ray {CT} scanner,''
\newblock {\em Measurement Science and Technology}, vol. 30, no. 3, pp. 035401, 2019.

\bibitem{Biguri2016tigre}
Ander Biguri, Manjit Dosanjh, Steven Hancock, and Manuchehr Soleimani,
\newblock ``{TIGRE}: a {MATLAB-GPU} toolbox for {CBCT} image reconstruction,''
\newblock {\em Biomedical Physics \& Engineering Express}, vol. 2, no. 5, pp. 055010, sep 2016.

\bibitem{Biguri2025tigrev3}
Ander Biguri, Tomoyuki Sadakane, Reuben Lindroos, Yi~Liu, Malena Sabaté~Landman, Yi~Du, Manasavee Lohvithee, Stefanie Kaser, Sepideh Hatamikia, Robert Bryll, Emilien Valat, Sarinrat Wonglee, Thomas Blumensath, and Carola-Bibiane Schönlieb,
\newblock ``{TIGRE v3}: Efficient and easy to use iterative computed tomographic reconstruction toolbox for real datasets,''
\newblock {\em Engineering Research Express}, vol. 7, no. 1, pp. 015011, mar 2025.

\end{thebibliography}

\end{document}